\documentclass{article}
\usepackage{ijcai23}

\usepackage{times}
\usepackage{soul}
\usepackage{url}
\usepackage[hidelinks]{hyperref}
\usepackage[utf8]{inputenc}
\usepackage[small]{caption}
\usepackage{graphicx}
\usepackage{amsmath}
\usepackage{amsthm}
\usepackage{booktabs}
\usepackage{algorithm}
\usepackage{algorithmic}
\usepackage[switch]{lineno}

\usepackage{placeins}

\usepackage{fancyhdr}
\title{
Transcending XAI Algorithm Boundaries through End-User-Inspired Design
}

\author{
Weina Jin$^1$
\and
Jianyu Fan$^1$
\and 
Diane Gromala$^1$
\and  \\
Philippe Pasquier$^1$
\and
Xiaoxiao Li$^2$
\and
Ghassan Hamarneh$^1$
\affiliations
$^1$Simon Fraser University\\%
$^2$The University of British Columbia
}

\begin{document}

\maketitle

\begin{abstract}
The boundaries of existing explainable artificial intelligence (XAI) algorithms are confined to problems grounded in technical users' demand for explainability.
This research paradigm disproportionately ignores the larger group of non-technical end users, who have a much higher demand for AI explanations in diverse explanation goals, such as making safer and better decisions and improving users' predicted outcomes.
Lacking explainability-focused functional support for end users may hinder the safe and accountable use of AI in high-stakes domains, such as healthcare, criminal justice, finance, and autonomous driving systems.
Built upon prior human factor analysis on end users' requirements for XAI, 
we identify and model four novel XAI technical problems covering the full spectrum from design to the evaluation of XAI algorithms, including edge-case-based reasoning, customizable counterfactual explanation, collapsible decision tree, and the verifiability metric to evaluate XAI utility. Based on these newly-identified research problems, we also discuss open problems in the technical development of user-centered XAI to inspire future research.
Our work bridges human-centered XAI with the technical XAI community, and calls for a new research paradigm on the technical development of user-centered XAI for the responsible use of AI in critical tasks.

{
}
\end{abstract}

\section{Introduction}

\begin{figure*}
    \centering
    \includegraphics[width=1\textwidth]{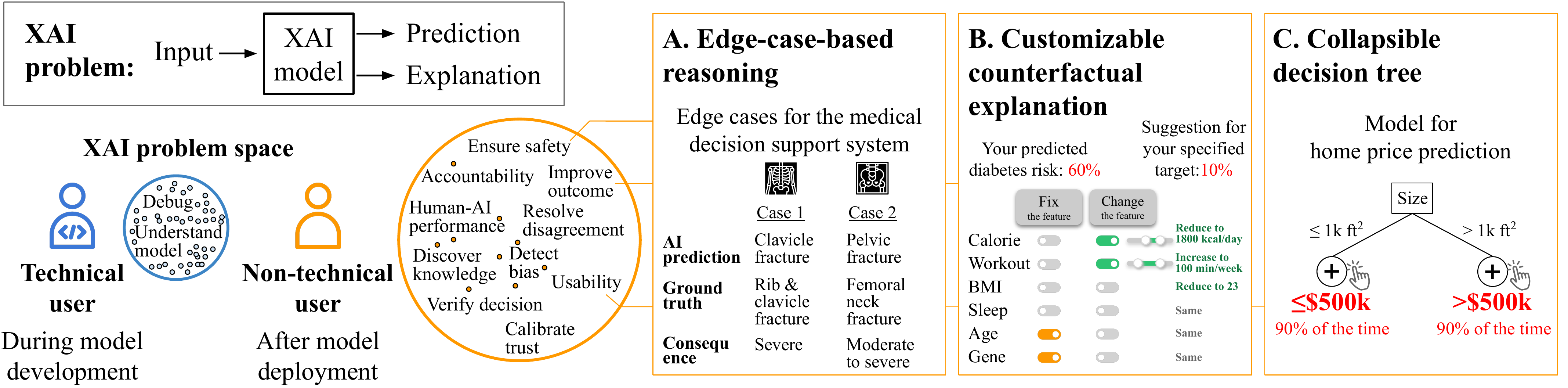}
    \caption{
The key ideas of this work are graphically highlighted. 
The general explainable AI (XAI) problem is to provide an explanation alongside the AI model's prediction. We visualize the XAI problem spaces with circles, including XAI problems that technical users or non-technical users are concerned about, respectively. Dots inside a circle visualize the density of technical solutions in the problem space. For technical users, their main demands for XAI are model debugging, understanding, and improvement during the model development phase. And the corresponding technical works are relatively saturated. In contrast, the non-technical users have a high demand for XAI for a variety of purposes, most of them related to high-stakes decisions during the longer usage period after model deployment. But the technical development for non-technical users to address their various XAI problems is disproportionately limited. 
Our paper reveals this research gap. To address it, 
we integrate users' requirements into the modeling of several novel XAI research problems and illustrate their application in critical tasks in panels A, B, and C. 
They address the XAI problems of ensuring safety, improving user's predicted outcome, and improving the usability of XAI techniques, respectively. 
    }
    \label{fig:abs}
\end{figure*}

The research development of explainable artificial intelligence (XAI) or interpretable machine learning (iML) algorithms has reached a steady state after the boom in this field several years ago
\cite{10.1007/978-3-030-65965-3_28}. 
Despite many XAI algorithms being proposed, it is unclear how to use them and evaluate their real-world impact. Lipton~\shortcite{Lipton2017} criticized this phenomenon as ``a surfeit of hammers, and no agreed-upon nails''.
This is partially due to the sole motivation that drives XAI algorithm development~\cite{miller2017explainable}: XAI algorithms are designed by AI people for AI people, 
with the explanation problems solely grounded in model debugging, understanding, and improvement.

Meanwhile, designing XAI algorithms for end users to support their critical tasks has high demand and significant social impacts, but receives disproportionately little attention in the current XAI research paradigm~\cite{https://doi.org/10.48550/arxiv.2206.01507}.
As AI becomes pervasive in high-stakes 
tasks -- such as in supporting end users' medical, legal, financial, or driving judgments -- making AI explainable to its end users is crucial to ensure the safe, ethical, and legal~\cite{gdpr} use of AI in critical domains.
The non-technical end users, or end users for short, do not have technical knowledge in AI, ML, or programming. They could be laypersons or domain experts, such as physicians, judges, drivers, bankers, and other critical decision-makers. Compared to technical users' demand for XAI towards model debugging, end users have more diverse explainability goals and needs.
Prior evaluations showed that directly applying existing XAI algorithms to end users' tasks failed to achieve the key function of explainability in detecting model biases~\cite{adebayo2022post}, or lacked key information on feature description to support clinical reasoning~\cite{JIN2023102684}. Furthermore, evaluations have shown that XAI can mislead users and create harmful consequences, such as worsening of physicians' performance due to misaligned XAI development objectives and under-specified users' requirements~\cite{Rethinking,10.1145/3375627.3375833,Jacobs2021}. 
These issues make designing XAI techniques for end users a more pressing and challenging research problem, one that needs to be addressed to ensure that AI and its explanation effectively support end users' critical decisions.
Designing XAI for end users also has a significant social impact: it makes AI more accessible to a larger group of users, who use AI to perform critical tasks. 
These unmet needs and significant social impact makes designing user-centered XAI algorithms a promising research direction, a direction that can potentially expand the problem space of XAI techniques.

Explainability is a ``fundamentally human-centered concept''~\cite{10.7551/mitpress/12186.003.0014}, yet current research trends on the modeling of XAI technical problems are not well motivated by the understanding of users' requirements. 
To expose the XAI technical community to human-centered perspectives,
in this work, 
backend by user study evidence on users' requirements for explanation goals and XAI algorithms (manifested by specific explanation forms)~\cite{euca}, we provide problem modeling on four novel XAI research problems: 
edge-case-based reasoning to help user understand AI capability and ensure safety, customizable counterfactual explanation to help user improve their predicted outcome, collapsible decision tree to improve the usability of decision tree, and the verifiability metric to evaluate XAI utility on helping user verify AI decisions (Fig.~\ref{fig:abs}). From the problem modeling, we further discuss open problems including how to learn and extract human-interpretable features from deep neural networks, how to optimize the form of explanation, and how to computationally evaluate and optimize for human-grounded properties of XAI algorithms. 
Our contributions are:
\textbf{(1)} At a high level, we highlight the important but technically ignored research topic of designing and evaluating XAI algorithms for user-centered properties. As the first transdisciplinary and bridge work that connect human-computer interaction (HCI) with AI community, we demonstrate the detailed process on how to carve user requirements into technical problems. 
\textbf{(2)} At the fine-grained level, we are the first to systematically model novel technical problems driven by users' requirements, covering the full spectrum from the design to the evaluation of human-centered XAI techniques. 

{

}

\section{Related work}
\textbf{End-user-inspired XAI techniques}
We summarize works that were inspired by end users' requirements for XAI.
Wang and Rudin~\shortcite{pmlr-v38-wang15a} proposed the falling rule list that can prioritize 
high-risk predictions for doctors to inspect, which 
aligns with human decision-making patterns in critical tasks. Biran and McKeown~\shortcite{ijcai2017-0202} used natural language to generate human-centered important feature explanations focused on domain knowledge and human reasoning. 
Adebayo et al.~\shortcite{adebayo2022post} conducted computational evaluation and a user study that identified post-hoc explanations as ineffective in helping end users detect unknown spurious patterns that a model relies on. Jin et al.~\shortcite{Jin_Li_Hamarneh_2022} proposed a computational evaluation metric grounded in physicians' clinical image interpretation patterns, and conducted a systematic evaluation 
on 16 saliency map algorithms.
Their results showed the examined algorithms were 
ineffective in helping clinical users appreciate the AI model decision process or gauge the decision quality. These works show the importance of and research gaps in end-user-oriented XAI algorithms.

\textbf{Human-centered XAI: design principle and user study}
The HCI community calls for human-centered XAI, which uses human-centered design methodologies to propose and assess XAI techniques based on the understanding of users and their tasks
~\cite{DBLP:journals/corr/abs-2110-10790,10.7551/mitpress/12186.003.0014}. 
General XAI design principles for user understanding are proposed based on psychology and social science~\cite{Miller2017,Wang2019,Liao2020}. 
Specific user studies were also conducted to understand end users' perceptions on various XAI techniques, such as 
similar and counterfactual examples~\cite{Cai2019}, 
rule-based explanations~\cite{10.1016/j.dss.2010.12.003}, 
feature attribution, typical, and counterfactual examples
~\cite{EVANS2022281}, and a set of 12 user-friendly explanation forms~\cite{euca}. 
These user studies on XAI requirement engineering and evaluation can serve as an important source of inspiration for the proposal of new XAI techniques. 
But there are rare transdisciplinary works that focus on informing technical development by user's requirements, and conducting user studies based on technical questions. 
To the best of our knowledge, we are the first to bridge AI research community with human-centered XAI by systematically modeling technical problems based on user understanding.

\textbf{Summary of the EUCA framework and the end-user-friendly explanation forms}\label{ef}
Our modeling of technical problems is grounded in the findings from the End-User-Centered XAI (EUCA) user study, which provides a comprehensive analysis of users' requirements for XAI techniques~\cite{euca}.
The EUCA framework summarizes existing explanation forms -- the abstracted structure of information generated by an XAI algorithm -- from the technical literature, filters the ones that do not require technical knowledge to interpret them, and names them the end-user-friendly explanation forms, including feature- (feature attribution, feature shape, feature interaction), example- (similar, typical, and counterfactual example), and rule-based explanations (decision rule and decision tree). 
EUCA also summarizes common explanation goals that motivate an end user to check explanations, such as ensuring safety, calibrating trust, improving predicted outcomes. 
EUCA conducted a user study with 32 laypersons using quantitative and qualitative methods, uncovering users' general requirements for these explanation forms and their interactions with different explanation goals, which serve as the main motivation for our proposal of user-centered XAI research problems.

\section{End-user-inspired modeling of XAI technical problems}\label{technical_problem}

We first set up the general explainability problem formulation: 
A predictive model $f$ maps an input $x$ in the input space $X$ to a target prediction $y_t$ in the output space $Y$,
i.e.: $y_t = f(x)$, with ground-truth prediction denoted as $\Tilde{y} \in Y$. An XAI algorithm $M$ generates explanations $e$ of how model $f$ arrives at its predictions.
A human-interpretable feature is denoted as $a$ in the human-interpretable feature space $A$.

Next,
we present the process of modeling novel XAI technical problems inspired by end users' requirements. 
For each problem, we present the motivation based on users' requirements for XAI from the EUCA user study findings. We analyze prior works, identify research gaps, formulate the technical problem, and sketch the potential technical solution.
We end by raising some challenges and open questions on end-user-oriented XAI algorithms.
\subsection{Edge-case-based reasoning}
\textbf{Motivation}
For AI-assisted decision support in high-stakes domains, such as healthcare, autonomous driving, and criminal justice, the consequence of making wrong decisions is substantial.  To
collaborate with AI and make better decisions,
users need to have a complete understanding of AI's capabilities, particularly when, why, and how AI works and fails. 
For example, in the domain of autonomous driving or healthcare, in addition to knowing how the AI reacts to typical or common cases, the driver or the doctor may want to know how the AI will react to some rare but significant situations: a driver may need to know how the car drives under bad weather conditions, during the night, or when a dog suddenly jumps onto the road~\cite{euca}; a doctor may need to know how the AI predicts a rare disease that appears similar to a common disease but has a fatal outcome if not recognized. Namely, in critical tasks and tasks related to safety issue, users may request to check how AI reacts to some edge cases to fully understand AI's capabilities.

\textbf{Research gap}
An intuitive way for users to understand AI's capabilities is to observe how AI behaves in different cases, and by accumulating such observations,
users can construct a mental model of AI (i.e., theory of mind~\cite{frith2005theory}) including its strengths and weaknesses. In fact, perturbation-based post-hoc XAI algorithms borrow such idea 
to understand, summarize, and explain any black-box ML model by perturbing the input and sampling the model's input-output pairs~\cite{JIN2023102009}. Fine-grained model performance metrics also summarize the model's performances under different scenarios. In addition to the summarized explanation model or the stated performance, users prefer to gain hands-on experience on AI model's behavior by inspecting individual cases, described in previous work as observed performance~\cite{Yin2019} and example-based explanation. In particular, the example-based explanation utilizes case-based reasoning -- a paradigm that emulates human learning and decision-making processes to solve new problems based on past solutions~\cite{rudin2022interpretable} -- to explain the AI model's behavior using its decisions for similar or prototypical examples in the past. 
However, the primary focus of the preceding approaches is to use major or common cases to explain AI's core capabilities of when and what the AI is good at, and are limited in their ability to explain the model's behavior beyond the common cases. 

Another line of research topics related to non-typical, edge, or failure cases includes anomaly detection, uncertainty estimation~\cite{10.5555/3045390.3045502}, learning to defer~\cite{10.5555/3524938.3525594}, long-tailed data learning~\cite{NIPS2017_147ebe63}, and out-of-distribution detection~\cite{https://doi.org/10.48550/arxiv.2211.16158}. These methods are primarily used for predictive tasks and have not been applied to explainability tasks to explain AI capabilities to users, particularly AI's shortcomings.

\textbf{Problem formulation}
Based on users' requirements to understand AI's capabilities and ensure safe decision-making through their inspection of cases-of-interest, we formulate the technical problem of edge-case-based reasoning (Fig.~\ref{fig:abs}-A), which provides a set of, or a summary of edge cases that possess important properties, such as having potentially negative consequences if incorrectly predicted. 

\textbf{Sketched solution} 
Given an AI model $f$, we aim to identify or construct its edge case set $\mathcal{E}$ that meet users' criteria of interest: 
\begin{align}\label{edge}
    \mathcal{E}(f, x_{opt}) = \{x_e \in X | C(f, x_e, x_{opt})\}
\end{align}
where $x_{opt}$ is an optional parameter denoting a query input, and $C$ is a logical formula that describes a conjunction of criteria of interest defining the instance $x_e$. The criteria of interest can be predefined or specified by users. A common $C$ that defines edge cases can be: 
\begin{align}\label{consq}
C(f, x_e, x_{opt}; r) =  
\begin{cases}
Risk(x_e) >  r \\
f(x_e) \neq  \Tilde{y} 
\end{cases}
\end{align}
where $\Tilde{y}$ is the ground truth prediction for the edge case $x_e$; $Risk$ is the consequence function that outputs a risk value if an instance is incorrectly predicted; and $r$ is the risk threshold. Eq.~\ref{consq} defines the edge case $x_e$ as an instance that is wrongly predicted, and its consequence of being wrongly predicted assessed by $Risk$ is above the threshold $r$.

In practice, the logic formula can be converted to multi-objective optimization, or to one scalarized optimization function with a weighted sum. 
If $x_{opt}$ is not provided, then $x_e$ is a global edge case that shows the overall behavior of $f$. Otherwise, $x_e$ is a local edge case related to $x_{opt}$. For example, by adding a constraint to $C$, $x_e$ can be a counterfactual example that mimics $x_{opt}$ in every aspect, except that $x_e$ has a much significant consequence. If the edge case set $\mathcal{E}$ is large, it can further be summarized using explanation models or performance metrics as stated above.

\subsection{Customizable counterfactual explanation}
\textbf{Motivation}
Counterfactual explanation is a common explanatory method used to explain the current outcome in contrast to a hypothetical outcome that would happen if some factors (counterfactual features) had changed~\cite{ijcai2019p876}. 
The three most common explanation goals for users to seek counterfactual explanations are: 
improving users' predicted outcomes; helping users differentiate similar instances; and improving user's own learning and problem-solving skills~\cite{euca}. Because explanations should be relevant enough to help users answer their questions for these explanation goals, users request counterfactual explanations to be able to receive user-defined constraints on feature type and range. For example, for users seeking advice from an explanation to reduce their health risks, the counterfactual explanation should be constrained to only changing features that users can control, such as lifestyle features. Explanations without such a constraint may include uncontrollable features, such as genetic factors, gender, and age, which are not helpful for users to achieve their explanation goal of improving health outcomes. Furthermore, users may have an aversion to counterfactual explanations that suggest changes to uncontrollable features~\cite{euca}. In addition to the constraint on feature type, users may request to set the range limit for the change of a counterfactual feature. For example, the change of the counterfactual features in lifestyle should be within a user-defined range (e.g., the maximum number of weekly exercise hours). With these constraints, users may also request to rank and choose from a set of alternative counterfactuals. 

\textbf{Research gap}
Although some existing work on optimization for counterfactual explanations includes constraints on sparse and controllable feature change~\cite{ijcai2019p876}, counterfactual explanation techniques are not designed with users' intended explanation goals in mind to fully support their corresponding requirements and inquiries. In fact, counterfactual explanations may not be appropriate in every usage scenario, and previous research has shown that they can cause confusion for some explanation goals, such as improving users' understanding of the model~\cite{Cai2019,euca}.
The design of counterfactual explanations should be built upon an understanding of users' intended goals in seeking a counterfactual explanation.
As stated above, users' most requested explanation goals for counterfactual explanations may not focus on the model itself, but on seeking suggestions for users' own improvement. 
However, this explanation goal is not adequately supported by existing counterfactual XAI techniques,
and we should further introduce more customizable characteristics to counterfactual explanation techniques.

\textbf{Problem formulation}
To help users achieve their intended explanation goals in seeking counterfactual explanations to improve their predicted outcomes, we formulate the problem of user-customizable counterfactual explanation, where users can customize the sets of changeable and unchangeable features in a counterfactual explanation, and can specify the range of feature change if needed (Fig.~\ref{fig:abs}-B). 
In the search or optimization for counterfactual examples, the set of unchangeable features cannot be changed, the set of changeable features must be changed, and the rest unspecified features may or may not be changed. Specifying the set of unchangeable features is to avoid users' aversion as stated above. 
The set of unchangeable and changeable features, and the range of feature change, can be pre-defined or specified by the user. If user's interactive specification of the explanation is supported, it will add another requirement on the speed of XAI algorithm to get real-time counterfactual explanations.

\textbf{Sketched solution} 
The problem of user-customizable counterfactual explanation can be tackled by adding additional constraints to the optimization of counterfactual examples/explanations. Specifically, following a similar notation to that of Eq.~\ref{edge}, 
the set of counterfactual explanations $\mathcal{N}$ can be defined as:
\begin{align}
    \mathcal{N}(f, x, y_c) = \{x_c \in X | C(f, x, x_c, y_c)\}
\end{align}
where $x_c$ is a counterfactual instance, $y_c$ is the designated counterfactual outcome that is different from the prediction $f(x)$ of the query input $x$. A vanilla counterfactual instance is defined via the following constraints to $C^0$: 
\begin{align}
C^0(f, x, x_c, y_c; \epsilon)= 
\begin{cases}
    f(x_c) = y_c \\
    dist(x, x_c) < \epsilon
\end{cases}
\end{align}
where $dist$ denotes the distance between two instances, and $\epsilon$ is a dissimilarity threshold. To construct user-customizable counterfactual explanation, we impose the following additional constraints to $C$ to regularize the search or optimization for the set of $\mathcal{N}$:
\begin{align}
C =  C^0 \wedge 
\begin{cases}
A_u \subset A\setminus \Delta A\\
A_c \subset  \Delta A \\
a_i \in [L_i, H_i],  \forall a_i \in A_c \\
\min |\Delta A|
\end{cases}
\end{align}
where $\Delta A$ denotes the set of all the features that has been changed in $x_c$, and $A\setminus \Delta A$ denotes the set of unchanged features in $x_c$; $A_u$ and $A_c$ are the set of unchangeable and changeable features, respectively; 
and $H_i$ and $L_i$ specify the upper and lower range of a feature $a_i$ in the changeable feature set $A_c$;  
if unspecified, the default range is $a_i$'s value range in the training dataset; lastly, we add the common constraint to get sparse counterfactual explanations with minimal feature changes.
With the constraint set $C$, the set of user-customizable counterfactual explanation $\mathcal{N}$ can be identified via a search or optimization method, depending on the specific AI model and task. 

\subsection{Collapsible decision tree}

\textbf{Motivation}
Decision tree and decision rule/set (DT \& DR) are inherently interpretable models. A key challenge for DT \& DR is to balance interpretability and prediction accuracy: the increase of nodes or rules makes the prediction more accurate, but also makes DT \& DR more difficult for users to interpret and understand its decision process. To reduce users' cognitive load interpreting the explanation while retaining the predictive accuracy, users suggested showing shallow decision levels by default with a rough range of prediction, and showing more features and deep levels with a gradually precise range of prediction on demand~\cite{euca}.

\textbf{Research gap}
Despite the research focus on sparse DT \& DR to improve their interpretability~\cite{rudin2022interpretable}, no work has addressed users' above requirement to support the interactive functionality and to balance accuracy and interpretability. Related research topics include hierarchical classification~\cite{silla2011survey} and using decision trees for clustering~\cite{liu2000clustering}, but they do not address the explainability problem.

\begin{figure}[!h]
    \centering
    \includegraphics[width=1\columnwidth]{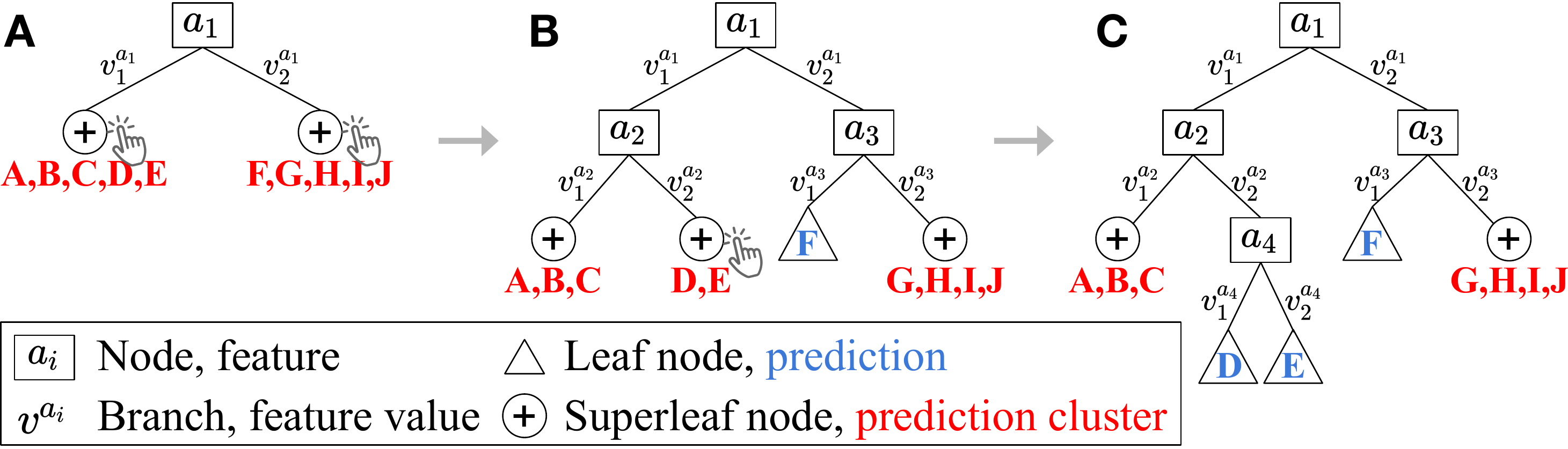}
    \caption{\textbf{Collapsible decision tree}. We change a traditional decision tree into a collapsible decision tree by truncating some non-leaf nodes and replacing them with superleaf nodes that summarize their descendant information as prediction clusters. The superleaf node, denoted by the $\oplus$ symbol, is expandable and collapsible interactively by clicking it, and the tree status after the superleaf expansion is shown in the subsequent panel. 
    The red and blue letters indicate the prediction cluster and individual prediction, respectively.}
    \label{fig:cdt}
\end{figure}

\textbf{Problem formulation}
To improve the usability of decision tree, we formulate the problem of sparse and collapsible decision tree that balances between concise and comprehensive explanation. 
In a collapsible decision tree (Fig.~\ref{fig:cdt}), the non-leaf nodes are substituted by superleaf nodes, which can be expanded or contracted via user's interaction by clicking it. 
By default, a collapsible decision tree is shown in a contracted state displaying only its shallow layers, as shown in Fig.~\ref{fig:cdt} panel A, and each superleaf node can be expanded, as shown in panels B and C, to allow users to dive deep into the decision process, especially to carefully inspect the branches where their query instance resides. This process is similar to the inspection of a file system or a table of content, in which semantically similar objects are grouped under the same subcategory in a hierarchical structure. This collapsible feature implicitly requires similar instances and their predictions in a decision tree to be clustered closely to each other under a superleaf node.

\textbf{Sketched solution} 
Built upon the existing sparsity constraint on decision tree (i.e., minimize the number of leaves)~\cite{rudin2022interpretable}, we add a new constraint to regularize the semantic distance between nodes belonging to a close common ancestor to be smaller than the distance of nodes from a remote common ancestor: 
\begin{align}\label{cdt}
    dist(o_i, o_j) < dist(o_i, o_k), \forall R(o_i,o_j) < R(o_i ,o_k)
\end{align}
where $R$ is the closeness of relationship between two nodes, measured by the number of branches by which two nodes $o_i$ and $o_j$ are connected to their common ancestor. The semantic distance $dist$ between two nodes can be measured by the average, minimum, or maximum distance between the pair of supporting instances (data instances that follow a rule path in the decision tree)  for each node. And the distance between the pair of data instances can be measured by the similarities of the data input feature $x$, the output prediction $y$, or the input-output tuple $(x,y)$. In this way, the collapsible decision tree is constructed similarly to a hierarchical clustering, and we regard each node in the decision tree as a cluster. Depending on the chosen decision tree algorithm, the constraint of Eq.~\ref{cdt} can be added as an additional criterion to measure the split in a decision tree.

\subsection{The verifiability metric to evaluate XAI utility}\label{formulation_eval}

\textbf{Motivation}
A fundamental user-centered aspect of XAI evaluation is to assess the utility of explanations to its users, i.e., whether the explanation can help users answer their questions and achieve their expected explanation goal~\cite{Rethinking}. In high-stakes domains, a common explanation goal is to verify AI decisions for complementary human-AI task performance, a performance that surpasses sole human or AI performance. After the deployment of AI and during its assistance to users on real-world tasks, due to the lack of ground-truth prediction, user's prior domain knowledge is indispensable information to discern potential model error and verify the correctness of model decision~\cite{chen2022machine}. The way users use their prior knowledge and explanation to judge AI decision quality or error is to assess the plausibility of explanation~\cite{jacovi-goldberg-2020-towards} (i.e.: how reasonable is the explanation compared to user's prior knowledge), and associate such plausibility assessment with AI decision quality~\cite{euca}.

\textbf{Research gap}
The landscape of XAI evaluation can be divided into two categories on whether the information of human knowledge or interpretation is involved in the evaluation: non-human-grounded evaluation and human-grounded evaluation. For non-human-grounded XAI evaluation that does not involve human interpretation, the evaluation objectives are on the technical or mathematical properties of explanation, such as truthfulness or faithfulness~\cite{jacovi-goldberg-2020-towards}, consistency or robustness. 
For human-grounded XAI evaluation, the evaluation objectives are on the utility of XAI, which could be the utility to the ultimate human-AI task performance~\cite{Jin2022.12.07.22282726}, and to various explanation subtasks (i.e., explanation goals) prior to the final decision-making task, such as human understandability, simulatability, usability, trustworthiness, the ability to help users verify model decisions, and identify model biases. 
These human-grounded evaluations heavily rely on human-subject studies and human-in-the-loop assessment, which are costly and inefficient to evaluate a batch of XAI algorithms on a dataset with sufficient sample size. Prior human-grounded evaluations do not fully utilize the explanation benchmark datasets that encode human knowledge and computational evaluation methods that can improve efficiency and automate the evaluation process.
Furthermore, existing XAI evaluation works rarely touch on the evaluation of XAI algorithms on their utility to verify model decisions, despite the fact that decision verification is a common motivation for users to check explanations, and optimizing this explanation goal can directly boost the complementary human-AI performance. To the best of our knowledge, there are only two works that conduct computational experiments to evaluate the utility of XAI in decision verification~\cite{JIN2023102684,NEURIPS2021_de043a5e}, but they did not define an evaluation metric to formalize the evaluation of XAI utility on verifying AI decisions.

Another line of work on human-grounded evaluation directly uses plausibility as the evaluation objective to assess the goodness of XAI algorithms. However, as Jin et al. analyzed, this approach is problematic and plausibility is a misleading objective for XAI algorithm evaluation, because improving plausibility is not relevant to the achievement of any explanation goals including model transparency, understandability, or trustworthiness~\cite{Rethinking}. 

\textbf{Metric formulation}
To evaluate the utility of XAI algorithms for AI model decision verification, we formulate the metric \textit{verifiability} $V$, which is based on the user study observation that humans usually associate the model's decision quality with the plausibility of its explanation. Different from prior evaluation metrics that problematically pursue highly plausible explanations, the verifiability metric regularizes the degree of explanation plausibility to be informative for model decision quality. 
This means an explanation can be optimized to be plausible or implausible, and its plausibility level is indicative of model decision quality, i.e., a highly plausible explanation is associated with an AI decision that is likely to be correct, and vice versa. The verifiability metric $V$ embodies the evaluation objective of informative plausibility in the clinical XAI guidelines~\cite{JIN2023102684}, and is constructed as follows:

\begin{equation}
V = \rho(\mathbf{p},\mathbf{q})
\end{equation}
where $\rho$ measures the correlation between the explanation plausibility $\mathbf{p}$ and the model decision certainty, confidence, reliability, or quality $\mathbf{q}$. $\mathbf{p}$ and $\mathbf{q}$ are vectors, usually aggregated from a number of test data, $\mathbf{p}=[p_1, \dots, p_n]^T$ and $\mathbf{q}=[q_1, \dots, q_n]^T$. The plausibility $p$ is also a computational metric, defined by the similarity between AI explanation $e$ and human explanation $k$: $p = Sim(e, k)$. In practice, the human explanation $k$ can be pre-annotated as a benchmark dataset, or generated by pre-trained models that encode human prior knowledge.

\textbf{Empirical evaluation} 
To assess the commonly-used feature attribution map algorithms on its utilities to verify AI decisions, we apply the verifiability metric to an AI-assisted medical image analysis task by analyzing raw data from the computational experiment and physician user study with data usage permission from the authors of~\cite{JIN2023102684,Jin2022.12.07.22282726}. The prior experiment built binary classification models to grade brain tumors on brain magnetic resonance image (MRI), generated feature attribution maps using different XAI algorithms, and asked neurosurgeons to assess the explanation quality and grade tumors with the assistance of AI and its explanation. We use the feature portion (the percentage of the explanation feature $e$ inside the tumor segmentation mask, which encodes the human knowledge $k$) to measure plausibility $p$ for an explanation, use the model softmax output probability of the \textit{ground-truth} class to represent the model decision quality $q$, and use the Spearman’s correlation coefficient as the correlation measure $\rho$. We list the top four XAI algorithms and their verifiability correlations in Table~\ref{tab:ver}, and the complete result is in the Appendix~\ref{a1}. The result shows that the verifiability for the best performing XAI algorithms have moderate correlations of around 0.5, indicating that they are not adequate to help users verify model decisions, because inspecting the reasonableness of explanations only reveals limited information on model decision quality. 

\begin{table}[!h]
\small
    \centering
    \begin{tabular}{c|c|c|c|c}
    \toprule
    XAI & Feature Ablation & Occlusion & Input$\times$Gradient & DeepLift \\ \hline
    $V$ & $.52\pm.06$
    & $.52\pm.08$
    & $.50\pm.08$
    & $.48\pm.10$\\
    \bottomrule
    \end{tabular}
    \caption{The top four XAI algorithms and their verifiability $V$ (M$\pm$SD). The experiment was repeated on five models that differed by the random state of parameter initialization.}
    \label{tab:ver}
\end{table}
We further validate the underlying assumption for the verifiability metric, that users rely on the explanation plausibility information to deduce the model decision reliability. On the prior user study data with 35 neurosurgeons each reading 25 MRIs~\cite{Jin2022.12.07.22282726}, we conduct an upper t-test to test whether the degree of explanation plausibility (rated by doctor on a 0-10 scale) is related to doctor's judgment on AI decision reliability. Since the user study did not directly measure doctor's judgement on decision reliability, we use the doctor's agreement with AI after seeing AI prediction and explanations as the indicator. The result validates the assumption and shows that for decisions that doctors agree with AI, doctors' mean plausibility rating for explanations is significantly higher (M=6.45, SD=2.82)
than decisions that doctors disagree with AI (M=3.82, SD=2.56), $t(130)=9.24$, $p$-value$=3.2\times10^{-16}$. The statistical summary and data visualization are in the Appendix~\ref{a2}. 

\subsection{Open problems on end-user-centered XAI}\label{open}
In addition to the above modeling of specific research problems, we raise the following open problems (OP) on the design and evaluation of end-user-oriented XAI algorithms.

\textbf{OP1: How to learn and extract human-interpretable features from deep neural networks?}

From the above problems of customizable counterfactual explanation and collapsible decision tree, we can see that human interpretable features $a$ are the central element to organize an explanation.
Features can be regarded as the common backbone of different explanation forms, and various forms of explanation differ by how they weave features into their individual reasoning chain to describe the model decision process:
feature-based explanations describe features $a_i, a_j, \dots \in A$ and their relationship with the outcome $y$ quantitatively;
example-based explanations present features $a_i, a_j, \dots$ within their context as instances $x_m, x_n, \dots$; and rule-based explanations organize features $a_i, a_j, \dots$ and outcomes $y$ by logic and conditional statements. 

For the state-of-the-art AI model -- deep neural networks (DNNs), since they usually receive high-dimensional input data such as images or texts, and their latent feature space and decision process are usually non-human-interpretable, it is an open problem on how can we effectively learn or extract human-interpretable features in DNN. 
Prior works have explicitly represented features in deep neural networks as feature attribution map~\cite{JIN2023102684}, concepts~\cite{Kim}, or typical examples~\cite{Chen,9577335}. Are there any other ways to represent human-interpretable features in DNN?

In addition, to further improve interpretability, how can we effectively use the human-interpretable features to explain the decision process in DNN? The decision process in DNN from feature representation to prediction can be constructed or interpreted using the reasoning structure from different explanation forms, such as feature attribution~\cite{Chen} and decision tree~\cite{9577335}. By this means, the new user-inspired XAI techniques, such as the proposed collapsible decision tree structure, can be integrated into DNN.
Using different feature representation forms with different reasoning structures of the decision process, there can be a combinatorial number of new XAI techniques.

\textbf{OP2: How to optimize the form of explanation?}

Since there can be a number of ways to combine different feature representation forms with different decision reasoning structures stated in OP1, the next open problem is how we can optimize for the different forms of explanation. Jin et al. suggest that the information corresponding to different explanation forms should fully support users' reasoning and decision process to improve understandability and avoid confirmation bias during user's interpretation of explanation~\cite{Rethinking,JIN2023102684}. The EUCA user study finds that participants tended to combine different explanation forms to fill an explanation goal, the preference of the choice and combination of explanation forms are highly individualized, and different explanation goals may have different combination patterns of the explanation forms~\cite{euca}.
These indicate that the optimization and proposal of new explanation forms need to consider the human reasoning and explanation interpretation process, and be validated via user study focusing on one or several explanation goals. It is also an open research problem to optimize for personalized and interactive explanations.

\textbf{OP3: How to computationally evaluate and optimize for human-grounded properties of XAI algorithms?}

One barrier to the development of user-centered XAI is the lack of computational evaluation and optimization pipelines. Current research on user-centered XAI heavily relies on human-subject experiments or human-in-the-loop assessment, which is non-differentiable, thus cannot be used to directly optimize XAI algorithms for explanation goals that end users are concerned about. Computational quantification can act as a proxy for human-related process to optimize DNN models for user-centered properties. The computational quantification and evaluation is meant to be a complementary evaluation to human-subject studies, and can be conducted prior to the formal user study. It is an open problem on how to abstract the human interpretation, reasoning, information synthesis, and decision process with explanation as quantifiable metrics.

In the formulation of the verifiability metric, we present a way to use human-annotated benchmark datasets or pre-trained models as a surrogate for human prior knowledge $k$, and replace the human explanation interpretation process with a computational assessment summarized by the plausibility metric $p$. To associate $p$ with the explanation goal to verify AI decisions, we also make the implicit assumption to simplify the human reasoning dynamics, which is supported by empirical user study data: humans make a rational decision to take or reject an AI recommendation based on their assessment on how reasonable the explanation is. These computational modeling and necessary assumptions on human behavior, together, enable us to use computational metrics to optimize XAI for a user-centered explanation goal. More works on understanding and modeling the dynamics of human-AI collaboration with explanation are needed to computationally optimize for user-centered XAI techniques~\cite{Bansal_Nushi_Kamar_Weld_Lasecki_Horvitz_2019}.

\section{Conclusion, limitation, and future work}
The current XAI research paradigm is mainly technical-user-centered, which disproportionately ignores non-technical end users' high demand for explainability in a variety of explanation goals. In this work, we propose a promising and impactful research direction of user-inspired XAI techniques, and conduct transdisciplinary research to identify and model novel XAI algorithms driven by XAI problems that end users are concerned about. 
As a bridge work that introduces human-centered perspectives to the AI community, our work is not intended to propose state-of-the-art technical solutions nor conduct detailed experiments for each individual problem modeled, as this is beyond the scope of a single research paper. Instead, we only suggest preliminary solutions and leave finding and evaluating superior solutions as open questions for future work. 
This work is not a comprehensive list of end-user-centered XAI problems. Rather, it aims to serve as a starting point to the systematic study of XAI problems inspired by end users' requirements. In other words,
the overarching goal of this work is to call for a new research paradigm of end-user-centered XAI. 
Future work should emphasize a synergy between AI technical researchers and  end users that hones down on specific XAI objectives that lead to wider trust and adoption of AI in a variety, especially task-critical, applications. 
This in turn may require conducting more targeted user studies focused on eliciting domain- or task-specific requirements with the potential to inspire the development of user-centered XAI algorithms.
{%
}

\section*{Ethical Statement}
We selected the ethics-approved EUCA user study as the basis for the technical problem formulation. This research does not involve human subjects. The reuse of user study data has gained the authors' permission. 
There are no ethical issues.

\appendix
\renewcommand{\thesection}{A\arabic{section}}
\onecolumn
\section*{Appendix}
\section{Evaluation result on verifiability}\label{a1}
\begin{table}[h]
    \centering
    \begin{tabular}{l|c}
    \toprule
        \textbf{XAI algorithm} & \textbf{Verifiability} \\ \hline
Feature Ablation & 0.52$\pm$0.06 \\ \hline
Occlusion~\cite{10.1007/978-3-319-10590-1_53,DBLP:conf/iclr/ZintgrafCAW17} & 0.52$\pm$0.08 \\ \hline
Input$\times$Gradient~\cite{shrikumar2017just} & 0.50$\pm$0.08 \\ \hline
DeepLift~\cite{10.5555/3305890.3306006} & 0.48$\pm$0.10 \\ \hline
Integrated Gradients~\cite{10.5555/3305890.3306024} & 0.47$\pm$0.08 \\ \hline
Gradient Shap~\cite{NIPS2017_8a20a862} & 0.45$\pm$0.09 \\ \hline
Shapley Value Sampling~\cite{CASTRO20091726} & 0.43$\pm$0.08 \\ \hline
Guided BackProp~\cite{springenberg2015striving} & 0.34$\pm$0.10 \\ \hline
Deconvolution~\cite{10.1007/978-3-319-10590-1_53} & 0.34$\pm$0.12 \\ \hline
Guided GradCAM~\cite{8237336} & 0.34$\pm$0.12 \\ \hline
Gradient~\cite{simonyan2014deep} & 0.33$\pm$0.10 \\ \hline
Smooth Grad~\cite{smilkov2017smoothgrad} & 0.32$\pm$0.13 \\ \hline
Kernel Shap~\cite{NIPS2017_8a20a862} & 0.30$\pm$0.15 \\ \hline
Lime~\cite{Ribeiro2016b} & 0.28$\pm$0.10 \\ \hline
Feature Permutation~\cite{JMLR:v20:18-760} & 0.15$\pm$0.18 \\ \hline
GradCAM~\cite{8237336} & 0.13$\pm$0.15 \\ 
\toprule
    \end{tabular}
    \caption{The evaluated 16 XAI algorithms and their verifiability $V$ (M$\pm$SD). The experiment was repeated on five models that differed by the random state of parameter initialization.}
    \label{tab:verify}
\end{table}
\FloatBarrier

\section{Testing for plausibility and users' decision verification}\label{a2}
\begin{table}[h]
    \centering
    \begin{tabular}{c|c|c|c|c|c|c|c}
    \toprule
     Group & Number of decisions   &  M$\pm$SD & Min & 25\% & Median & 75\% & Max \\ \hline
     Agree &   649 & 6.45$\pm$2.82 & 0 & 5 & 7 & 9 & 10 \\ \hline
     Disagree & 95 & 3.82$\pm$2.56& 0 &2 &4 &5.5 & 10\\
     \bottomrule 
    \end{tabular}
    \caption{Statistical summary of physicians' assessment of explanation plausibility for two decision groups, regarding whether doctors' final decisions agree or disagree with AI's prediction. Doctors rated the explanation plausibility on a 0–10 scale.}
    \label{tab:plausibility_agreement}
\end{table}
\FloatBarrier

\begin{figure}[!h]
    \centering
    \includegraphics[width=0.8\textwidth]{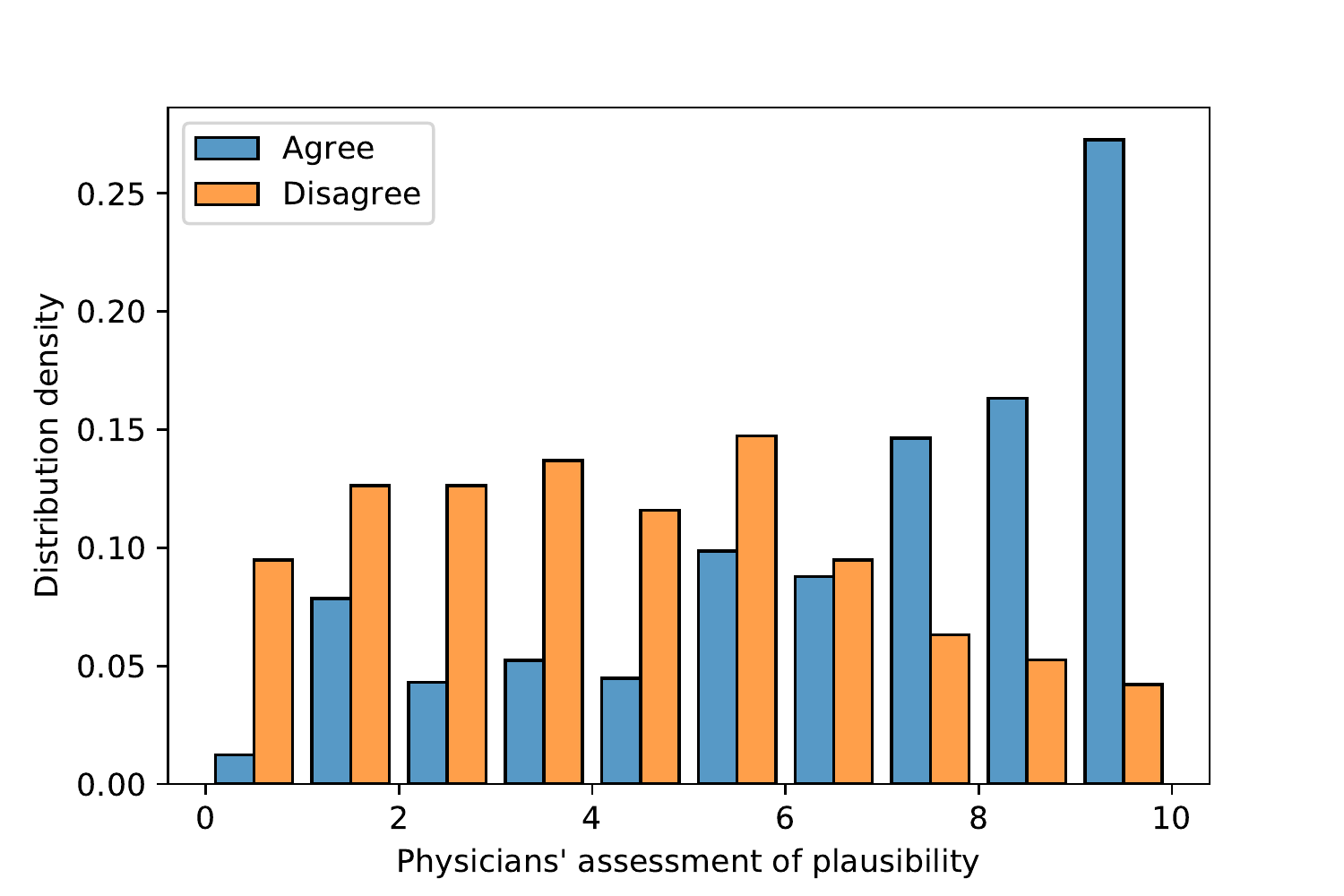}
    \caption{Distribution histogram of physicians' assessment of explanation plausibility. Physicians' assessment of the explanation plausibility is measured by their responses to the following question on a 0-10 scale: ``How closely does the highlighted area of the color map match with your clinical judgment?''  
The blue (left) and orange (right) bars are the distributions of groups when doctors' final decisions agree or disagree with AI's prediction, respectively. Since the data is imbalanced for the two groups, the histograms visualize for the relative distribution for each group.}
    \label{fig:plausibility_agreement}
\end{figure}
\FloatBarrier

\twocolumn
\bibliographystyle{named}
\bibliography{xai}

\end{document}